%% file: main.tex
\documentclass[10pt,twocolumn,letterpaper]{article}

\usepackage{cvpr}              

\input{preamble}

\definecolor{cvprblue}{rgb}{0.21,0.49,0.74}
\usepackage[pagebackref,breaklinks,colorlinks,citecolor=cvprblue]{hyperref}

\def\paperID{68} 
\def\confName{CVPR}
\def\confYear{2024}

\title{VWise: A novel benchmark for evaluating scene classification for vehicular applications}

\author{Pedro Azevedo$^{1}$ \quad Emanuella Ara\'ujo$^{1}$ \quad Gabriel Pierre$^{1}$ \quad Willams de Lima$^{1}$ \\ \quad João Marcelo Teixeira$^{1,2}$ \quad Valter Ferreira$^{3}$ \quad Roberto Jones$^{4}$ \quad Veronica Teichrieb$^{1}$ \vspace{0.3em} \\
{\normalsize $^1$Voxar Labs, Centro de Informática, Universidade Federal de Pernambuco} \quad \\
{\normalsize $^2$Departamento de Eletrônica e Sistemas, Universidade Federal de Pernambuco} \quad \\
{\normalsize $^3$Volkswagen Trucks and Bus} \quad
{\normalsize $^4$Eyeflow.AI}
}
\begin{document}

\twocolumn[{%
\renewcommand\twocolumn[1][]{#1}%
\maketitle
\begin{center}

    \centering
    \captionsetup{type=figure}
    
    \begin{subfigure}{0.3\linewidth}
        \includegraphics[width=\linewidth]{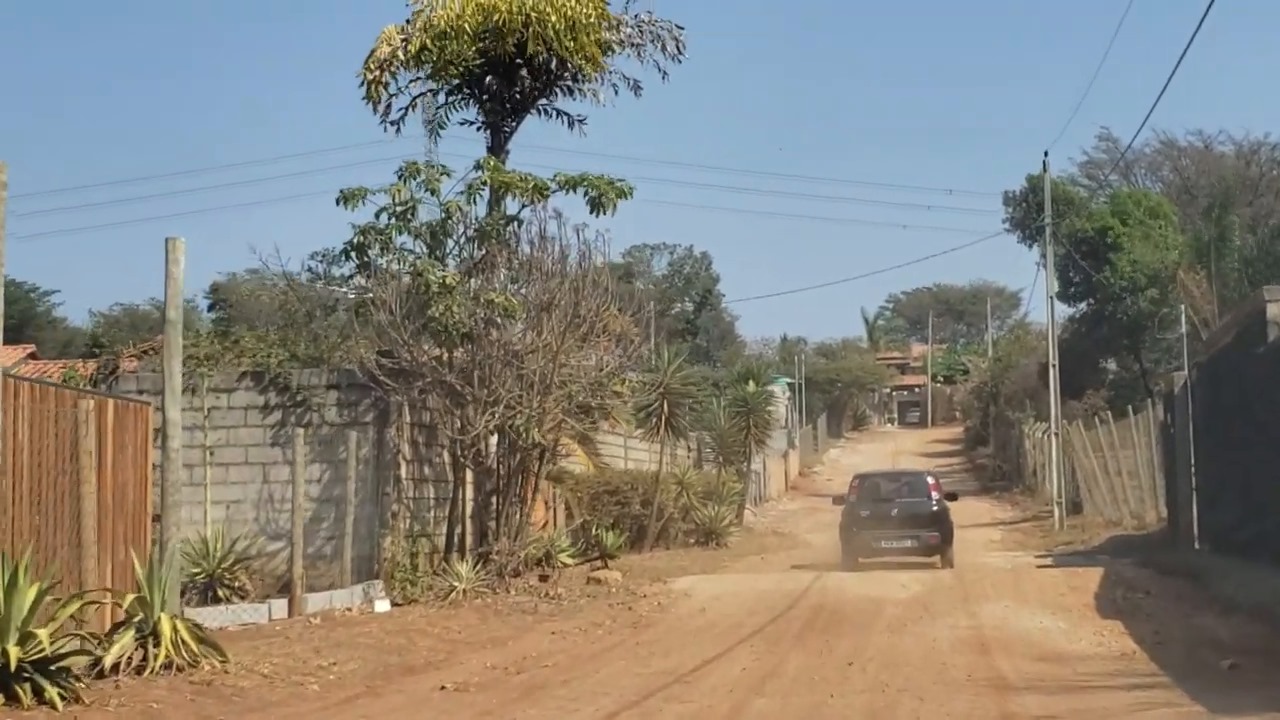}
        \caption{Dirt road}
    \end{subfigure}
\begin{subfigure}{0.3\linewidth}
        \includegraphics[width=\linewidth]{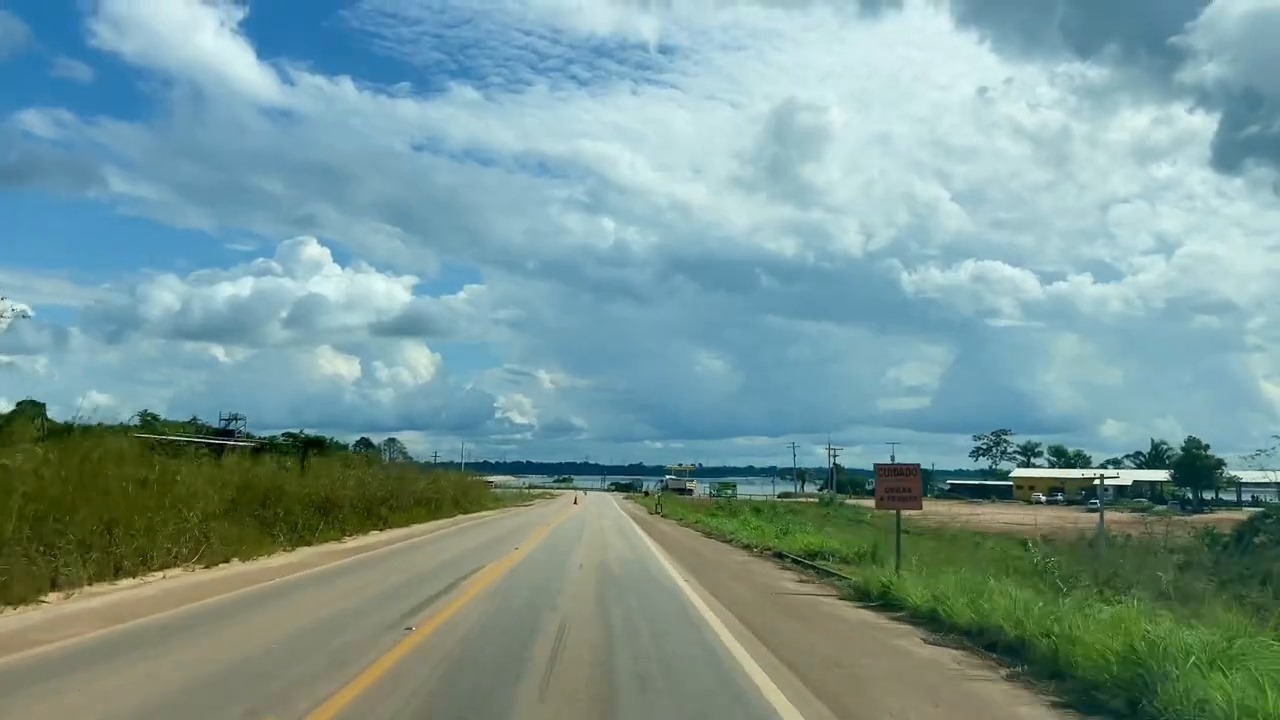}
        \caption{Highway}
    \end{subfigure}
    \begin{subfigure}{0.3\linewidth}
        \includegraphics[width=\linewidth]{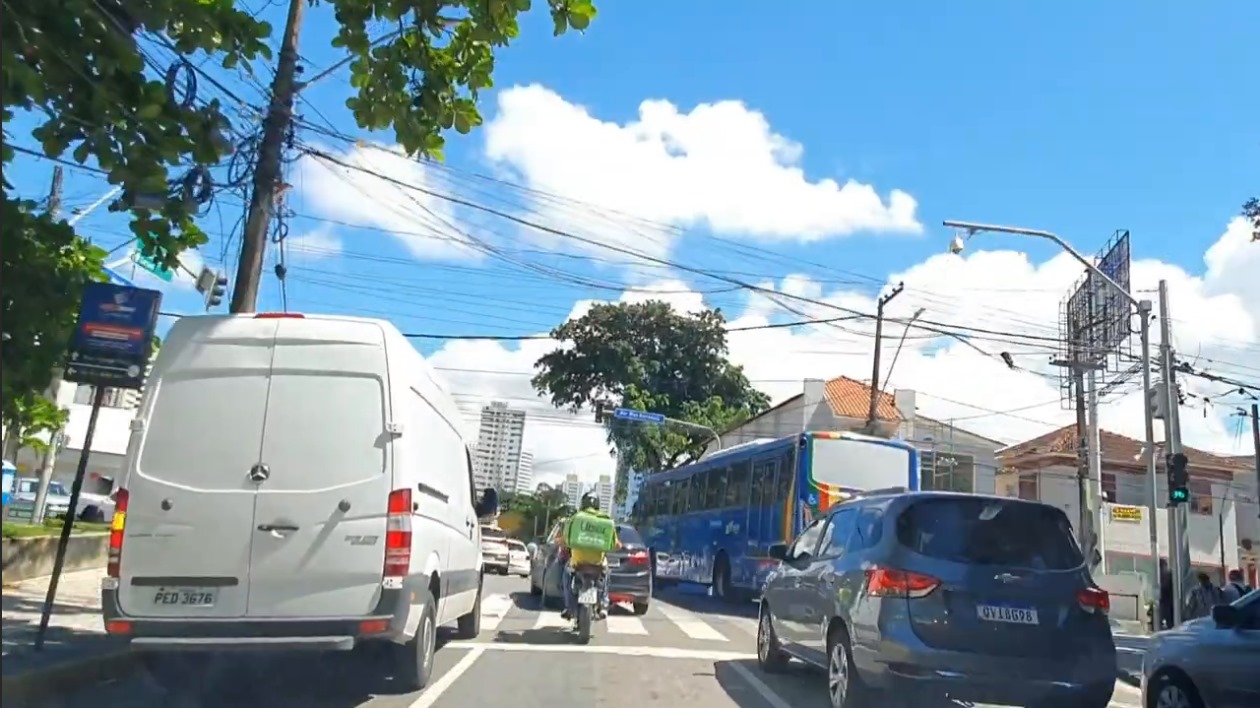}
        \caption{High-density road}
    \end{subfigure}
        \begin{subfigure}{0.3\linewidth}
        \includegraphics[width=\linewidth]{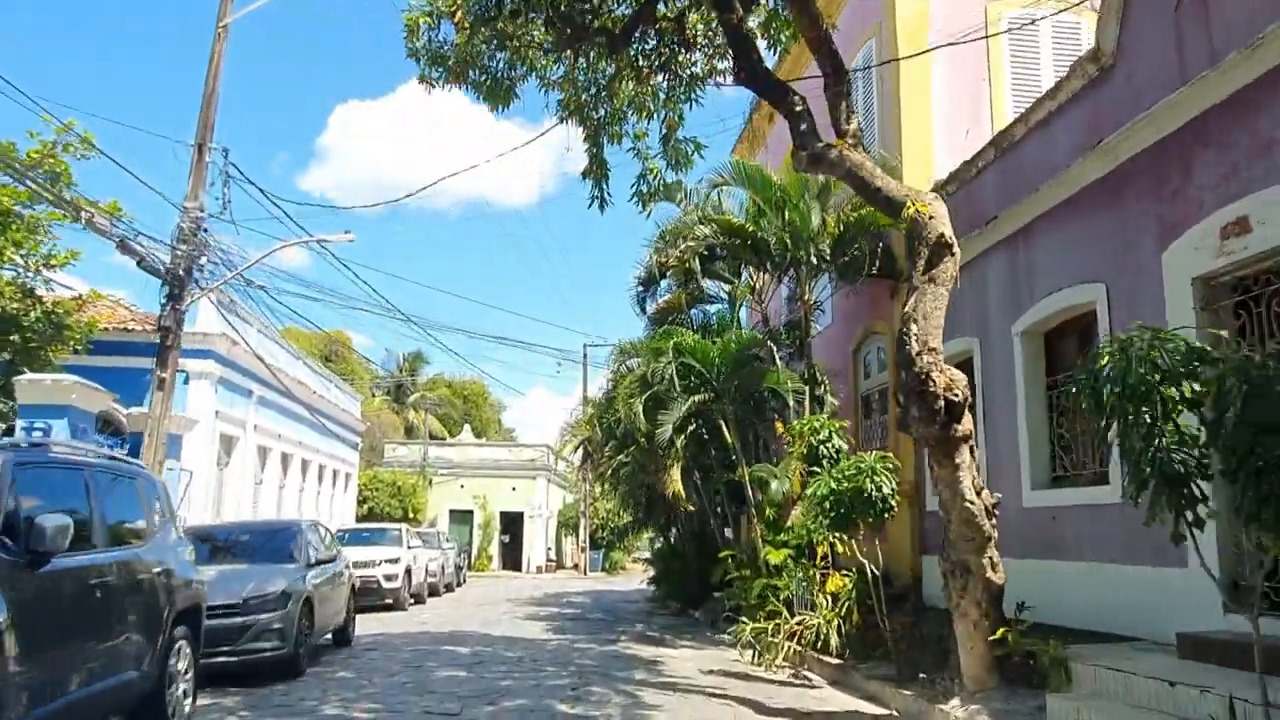}
        \caption{Low-density road}
    \end{subfigure}
        \begin{subfigure}{0.3\linewidth}
        \includegraphics[width=\linewidth]{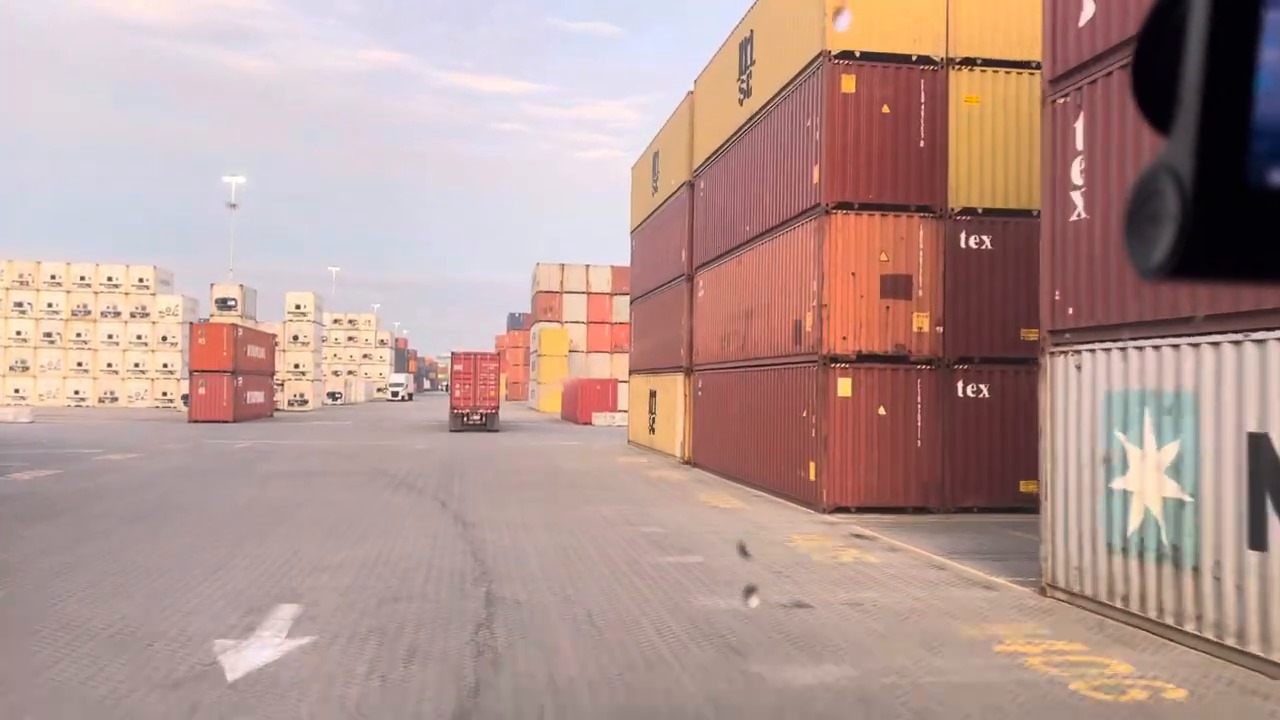}
        \caption{Port or loading dock}
    \end{subfigure}
        \begin{subfigure}{0.3\linewidth}
        \includegraphics[width=\linewidth]{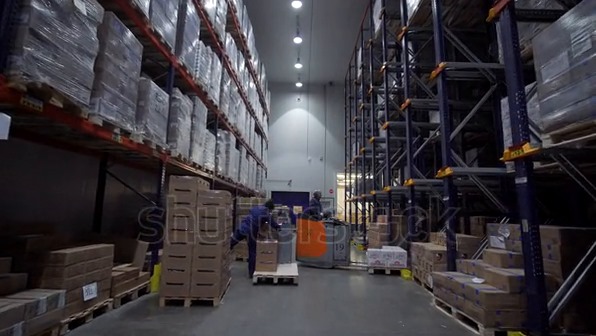}
        \caption{Factory access area}
    \end{subfigure}
    \setcounter{figure}{0} 
    \captionof{figure}{An overview of the VWise dataset, containing six classes from urban, industrial, and rural settings.}
\end{center}%
}]
\input{sec/0_abstract}    
\input{sec/1_intro}
\input{sec/2_related_work}
\input{sec/3_dataset}
\input{sec/4_results}
\input{sec/5_conclusion}
{
    \small
    \bibliographystyle{ieeenat_fullname}
    \bibliography{main}
}


\end{document}

%% file: preamble.tex
\usepackage[dvipsnames]{xcolor}


%% file: sec/0_abstract.tex
\begin{abstract}
\vspace{-1cm}
Current datasets for vehicular applications are mostly collected in North America or Europe. Models trained or evaluated on these datasets might suffer from geographical bias when deployed in other regions. Specifically, for scene classification, a highway in a Latin American country differs drastically from an Autobahn, for example, both in design and maintenance levels. We propose VWise, a novel benchmark for road-type classification and scene classification tasks, in addition to tasks focused on external contexts related to vehicular applications in LatAm. We collected over 520 video clips covering diverse urban and rural environments across Latin American countries, annotated with six classes of road types. We also evaluated several state-of-the-art classification models in baseline experiments, obtaining over 84\% accuracy. With this dataset, we aim to enhance research on vehicular tasks in Latin America.
\end{abstract}

%% file: sec/1_intro.tex
\section{Introduction}
\label{sec:intro}

With the current results in fields related to vehicular applications, such as autonomous driving and driver-assistance systems, interest in these applications has been rising over the past years. Overall, new applications are made available and being deployed on multiple vehicles in regions such as North America or Europe \cite{yin2017use}, such as the detection of pedestrians, vehicles, and traffic lights \cite{wang2021you}, the identification of pathways in streets and highways \cite{honda2024clrernet}, as well as evaluating their performance under different climatic conditions \cite{liu2022image}.

Multiple datasets have been proposed to facilitate the arrival of these technologies; however, there is a clear gap in deploying models trained and evaluated in these datasets to other regions of the world, especially Latin America, which is caused mainly by the different contexts between these regions. A clear example of this difference is how roads in Latin America are mostly unmaintained, while roads in North America and Europe have a higher maintenance rate. Therefore, collecting data and building datasets that would expand the geographical representation of these studies could lead to higher performance in these regions.

In light of this gap, we propose VWise, a novel dataset mainly focused on external context classification tasks, especially road classification, collected in various regions in Latin America. We expect that VWise will support training and evaluating state-of-the-art models, enriching research in tasks related to vehicular applications, such as autonomous driving or advanced driving assistance systems, and opening paths for deployment in this region.

%% file: sec/2_related_work.tex
\section{Related works}
\label{sec:formatting}


In automotive research, datasets play a pivotal role, dividing into categories such as those utilizing static cameras, exemplified by CityFlow \cite{tang2019cityflow}, which gathers images from traffic security cameras and annotates them. Another significant category includes datasets with onboard cameras like CityScape \cite{cordts2016cityscapes}, offering driver’s perspective views of streets, pedestrians, and road obstacles.

Focusing on vehicle onboard-captured images, we distinguish between two main categories: Indoors, showcasing the vehicle's interior, and Outdoors, displaying the vehicle's exterior. Indoor datasets observe actions within vehicles, focusing on drivers' or passengers' emotions, attention, and situational awareness. Examples like 100-Driver \cite{wang2023100} and MDAD \cite{jegham2019mdad} examine driver distraction with a focus on posture, while others delve into distraction based on head or eye movements \cite{schwarz2017driveahead, cyganek2014hybrid}.

For scene recognition, capturing onboard outdoor images is essential. Public datasets geared towards autonomous driving, such as KITTI \cite{geiger2012we} and Cityscape \cite{cordts2016cityscapes}, primarily focus on traffic environment perception without providing specific road type annotations. While CityScape \cite{cordts2016cityscapes} and KITTI \cite{geiger2012we} offer a populous dataset with detailed descriptions of buildings, objects, vehicles, and humans, they lack diversity regarding scene availability. Similarly, nuScene \cite{caesar2020nuscenes}, Waymo Open \cite{sun2020scalability}, ApolloScape \cite{huang2018apolloscape}, and ONCE \cite{mao2021one}, although large-scale and real-world, supporting 3D object detection, tracking, and activity predictions, share the limitation of not situating the vehicle in a specific scenario, like a rural dirt road or a metropolitan avenue, potentially leading to inaccurate activity predictions.

Public datasets like the Road Surface Classification Dataset (RSCD) \cite{zhao2023comprehensive} label certain road features but fail to specify the type of road the vehicle is on. Moreover, these data have a noticeable geographic concentration in North America and Europe \cite{yin2017use}. To address these gaps, we propose a new extensive dataset for classifying images and videos by road types to assist driving in various scenarios, focusing on Latin America. This effort is designed to overcome existing limitations by offering a resource that enhances the development and applicability of autonomous vehicle technologies across diverse geographical and situational contexts.

%% file: sec/3_dataset.tex
\section{The VWise Dataset}

\begin{figure}[h]
  \centering
  \begin{subfigure}{0.25\linewidth}
    \centering
    \includegraphics[width=\linewidth]{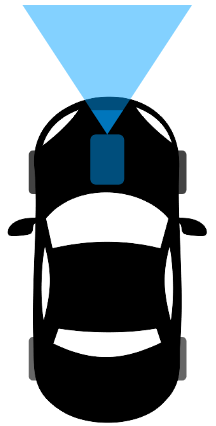}
    \caption{}
  \end{subfigure}
  \begin{subfigure}{0.5\linewidth}
    \centering
    \includegraphics[width=\linewidth]{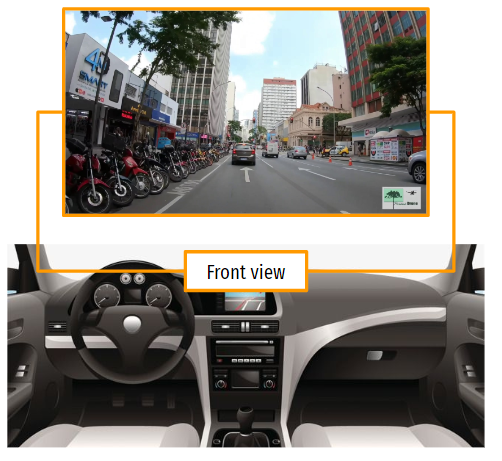}
    \caption{}
  \end{subfigure}
  \caption{Desired camera setup for VWise. A point-of-view camera setup allows us to have a similar view to the driver.}
  \label{fig:1}
\end{figure}

We collect the VWise dataset based on public access videos from YouTube and stock video libraries to tackle the lack of geographic representativeness in today's datasets. Queries related to countries (e.g., Brazil, Colombia, Peru...) and cities (e.g., Quito, Recife, São Paulo...) were combined with queries that would filter down to videos from point-of-view camera placements, as we exemplify in \autoref{fig:1}. Although this seems a naive approach, it allows us to quickly capture data from diverse regions of Latin America. The lack of quality in some videos, which are related to different cameras used in recordings (from GoPros to smartphones), actually contributes to a diverse dataset.

We have selected a diverse and comprehensive representation of urban and rural environments in multiple regions of Latin America, related to maintenance levels, suburban regions, or downtown regions, for example. We selected a total of 45 videos, from which we cropped 521 video clips, usually containing 10 seconds. We also processed a static version of the dataset for tasks that use images as input, generating a total of 4,500 images by sampling these videos. The dataset is divided into \textit{train}, \textit{test}, and \textit{val} splits following the well-accepted 70-20-10 split. For more details, please check the project page here: \url{https://vwise-dataset.github.io/}.

\subsection{Data annotation}
The data annotation process involves four students researching vehicular applications. Two are undergraduates; one is a master's student, and the other is a doctoral candidate, all from computer science and with experience in deep learning applications. The annotation process was performed blindly and independently. We have also employed CLIP \cite{radford2021learning} to enhance the efficiency of annotations. We have classified the data into six classes, as shown on \autoref{tab:labelDataset}.

\input{sec/label_dataset}

First, we generated descriptions from the list of classes to serve as input for CLIP, which bulk annotated the entire dataset. After this, multiple annotators verified the labels attributed by CLIP and attributed a new label based on a majority voting rule. Overall, we have seen that based on its zero-shot capabilities, CLIP has annotated most videos with significant accuracy.

\subsection{Data curation}
We have assessed our dataset using common data curation tools such as Cleanlab\footnote{Available at \url{https://cleanlab.ai/}} and Encord Active\footnote{Available at \url{https://github.com/encord-team/encord-active}} to check for the distribution of various image metrics such as area, ratio, brightness and distribution of RGB values to fix outliers and check if more data should be collected to distribute the dataset better. Overall, we have performed several iterations of checking the data and collecting more videos to reach a better data distribution.






%% file: sec/label_dataset.tex
\begin{table*}[h!]
  \centering
  \footnotesize
  \begin{tabular}{@{}ll r@{}}
    \toprule
     \textbf{Class} & \textbf{Description} & \textbf{Videos} \\
    \midrule
    Low-density urban street & Streets with low vehicle and pedestrian flow (residential areas and suburbs). & 97 \\
    High-density urban avenue & Avenues with high traffic density, vehicles, pedestrians, and traffic sign complexity. & 59 \\
    Highway & Fast tracks and highways, focusing on high-speed situations. & 124 \\
    Factory interior & Internal factory scenes, including assembly lines, storage, and logistics areas. & 24 \\
    Port or dock area & Port areas with intense loading and unloading activities, ships, and containers. & 86 \\
    Dirt or service road & Dirt or service roads, challenging navigation on irregular and less structured terrain. & 131 \\
    \bottomrule
  \end{tabular}
  \caption{Dataset classes, with their descriptions and number of videos in each class.}
  \label{tab:labelDataset}
\end{table*}

%% file: sec/4_results.tex
\section{Experimental baseline}
We have also evaluated several state-of-the-art techniques in our dataset to establish an experimental baseline. This experimentation allows other researchers to have a base for comparison in the future. In this evaluation, we set a baseline for the task of road-type classification.

\subsection{Image classification} We have selected common image classification architectures, namely EfficientNet-V2 \cite{tan2021efficientnetv2}, MobileNet-V3 \cite{howard2019searching}, and Vision Transformers (ViT-B-16) \cite{dosovitskiy2020image}, for the task of road-type classification.

\textbf{Implementation details.} We have used public implementations of these models as available for the PyTorch framework. They were trained on a single RTX 3050 6GB Laptop GPU using CrossEntropy and SGD with a learning rate of \(0.001\). 

\subsection{Zero-shot classification}
We have deployed two models for the task of zero-shot classification: CLIP \cite{radford2021learning} for images and X-CLIP \cite{ni2022expanding} for videos. In this case, as suggested by the good practices in the field, we have used the validation set of our dataset to learn a set of input prompts that would lead to a better result. We show the input prompts for each class in \autoref{tab:prompt_engineering}. We do not discuss the implementation details of these models since we follow the default implementation's training procedures.

\input{sec/input_prompt}

\subsection{Results and discussion}
For the image classification models, MobileNet-V3 has displayed the higher accuracy of the three models, with an accuracy of 95.02\%. EfficientNet-v2 was ranked second, with 91.16\% accuracy, and ViT-B-16 ranked third with 88.40\%. We show the results from this quantitative evaluation in \autoref{tab:image_classification}. Overall, we can see that the three architectures perform very well on this task and can learn strong representations from this data.

\begin{table}[ht]
  \centering
  \begin{tabular}{@{}lc@{}}
    \toprule
    Architecture & Accuracy (\%) \\
    \midrule
    VisionTransformer (Vit-b-16) & 88.40 \\
    EfficientNet-V2 & 91.16 \\
    \textbf{MobileNet-V3} & \textbf{95.02} \\
    \bottomrule
  \end{tabular}
  \caption{Baseline results on common image classification architectures, as trained and evaluated on VWise.}
  \label{tab:image_classification}
\end{table}

For zero-shot classification, CLIP and X-CLIP have reached similar accuracy levels; it is important to remember that CLIP uses images and inputs, while X-CLIP expects videos. For this evaluation, CLIP has reached an accuracy of 84.70\% and X-CLIP, 86.52\%, as we show in \autoref{tab:results-zeroshot-will}.

\begin{table}[ht]
  \centering
  \begin{tabular}{@{}lc@{}}
    \toprule
    Architecture & Accuracy (\%) \\
    \midrule
    CLIP & 84.70 \\
    X-CLIP & 86.58 \\
    \bottomrule
  \end{tabular}
  \caption{Baseline results on common zero-shot classification architectures, as trained and evaluated on VWise.}
  \label{tab:results-zeroshot-will}
\end{table}

%% file: sec/input_prompt.tex
\begin{table}[ht]
  \centering
  \begin{tabular}{@{}p{3.8cm}p{4cm}@{}}
    \toprule
    \textbf{Class} & \textbf{Input prompt} \\
    \midrule
    Dirt or service road & A rural dirt road or service road \\
    Factory interior & Factory interior \\
    High-density urban avenue & High-density urban avenue \\
    Highway & A highway suitable for road trips \\
    Low-density urban street & Low-density small urban road \\
    Port or dock area & Dock or port area \\
    \bottomrule
  \end{tabular}
  \caption{Prompts used for input in CLIP and X-CLIP.}
  \label{tab:prompt_engineering}
\end{table}

%% file: sec/5_conclusion.tex
\section{Conclusion}
In this work, we propose VWise, a novel dataset collected in Latin America for multiple vehicular tasks but focused on road-type classification. With this dataset, we expect to open paths for deploying automotive systems in this region since, nowadays, these models are mostly trained and evaluated on datasets sampled in North America or Europe. We also perform a baseline study that will serve as the basis for comparison for future works that will use VWise.

\subsection{Future works}
For future works, we plan on enhancing VWise with further annotations related to context to allow other applications to leverage it. Examples of annotations include ground-truth levels of road maintenance quality, ground-truth lane annotations, and other features related to the road surface. We also plan to study ambiguity; it is common to have highways acting as avenues within cities in countries such as Brazil. In this case, should these samples be annotated as highways or high-density roads? We have already pinned several cases of ambiguity, and we plan on tackling these cases before the final version of the dataset to allow for a more robust set of annotations. VWise is also the first step in a research pipeline that is planned within our research group. We plan on using this dataset to develop and deploy vehicular applications that make use of road-type classification.



%% file: main.bbl
\begin{thebibliography}{21}
\providecommand{\natexlab}[1]{#1}
\providecommand{\url}[1]{\texttt{#1}}
\expandafter\ifx\csname urlstyle\endcsname\relax
  \providecommand{\doi}[1]{doi: #1}\else
  \providecommand{\doi}{doi: \begingroup \urlstyle{rm}\Url}\fi

\bibitem[Caesar et~al.(2020)Caesar, Bankiti, Lang, Vora, Liong, Xu, Krishnan, Pan, Baldan, and Beijbom]{caesar2020nuscenes}
Holger Caesar, Varun Bankiti, Alex~H Lang, Sourabh Vora, Venice~Erin Liong, Qiang Xu, Anush Krishnan, Yu Pan, Giancarlo Baldan, and Oscar Beijbom.
\newblock nuscenes: A multimodal dataset for autonomous driving.
\newblock In \emph{Proceedings of the IEEE/CVF conference on computer vision and pattern recognition}, pages 11621--11631, 2020.

\bibitem[Cordts et~al.(2016)Cordts, Omran, Ramos, Rehfeld, Enzweiler, Benenson, Franke, Roth, and Schiele]{cordts2016cityscapes}
Marius Cordts, Mohamed Omran, Sebastian Ramos, Timo Rehfeld, Markus Enzweiler, Rodrigo Benenson, Uwe Franke, Stefan Roth, and Bernt Schiele.
\newblock The cityscapes dataset for semantic urban scene understanding.
\newblock In \emph{Proceedings of the IEEE conference on computer vision and pattern recognition}, pages 3213--3223, 2016.

\bibitem[Cyganek and Gruszczy{\'n}ski(2014)]{cyganek2014hybrid}
Bogus{\l}aw Cyganek and S{\l}awomir Gruszczy{\'n}ski.
\newblock Hybrid computer vision system for drivers' eye recognition and fatigue monitoring.
\newblock \emph{Neurocomputing}, 126:\penalty0 78--94, 2014.

\bibitem[Dosovitskiy et~al.(2020)Dosovitskiy, Beyer, Kolesnikov, Weissenborn, Zhai, Unterthiner, Dehghani, Minderer, Heigold, Gelly, et~al.]{dosovitskiy2020image}
Alexey Dosovitskiy, Lucas Beyer, Alexander Kolesnikov, Dirk Weissenborn, Xiaohua Zhai, Thomas Unterthiner, Mostafa Dehghani, Matthias Minderer, Georg Heigold, Sylvain Gelly, et~al.
\newblock An image is worth 16x16 words: Transformers for image recognition at scale.
\newblock \emph{arXiv preprint arXiv:2010.11929}, 2020.

\bibitem[Geiger et~al.(2012)Geiger, Lenz, and Urtasun]{geiger2012we}
Andreas Geiger, Philip Lenz, and Raquel Urtasun.
\newblock Are we ready for autonomous driving? the kitti vision benchmark suite.
\newblock In \emph{2012 IEEE conference on computer vision and pattern recognition}, pages 3354--3361. IEEE, 2012.

\bibitem[Honda and Uchida(2024)]{honda2024clrernet}
Hiroto Honda and Yusuke Uchida.
\newblock Clrernet: improving confidence of lane detection with laneiou.
\newblock In \emph{Proceedings of the IEEE/CVF Winter Conference on Applications of Computer Vision}, pages 1176--1185, 2024.

\bibitem[Howard et~al.(2019)Howard, Sandler, Chu, Chen, Chen, Tan, Wang, Zhu, Pang, Vasudevan, et~al.]{howard2019searching}
Andrew Howard, Mark Sandler, Grace Chu, Liang-Chieh Chen, Bo Chen, Mingxing Tan, Weijun Wang, Yukun Zhu, Ruoming Pang, Vijay Vasudevan, et~al.
\newblock Searching for mobilenetv3.
\newblock In \emph{Proceedings of the IEEE/CVF international conference on computer vision}, pages 1314--1324, 2019.

\bibitem[Huang et~al.(2018)Huang, Cheng, Geng, Cao, Zhou, Wang, Lin, and Yang]{huang2018apolloscape}
Xinyu Huang, Xinjing Cheng, Qichuan Geng, Binbin Cao, Dingfu Zhou, Peng Wang, Yuanqing Lin, and Ruigang Yang.
\newblock The apolloscape dataset for autonomous driving.
\newblock In \emph{Proceedings of the IEEE conference on computer vision and pattern recognition workshops}, pages 954--960, 2018.

\bibitem[Jegham et~al.(2019)Jegham, Ben~Khalifa, Alouani, and Mahjoub]{jegham2019mdad}
Imen Jegham, Anouar Ben~Khalifa, Ihsen Alouani, and Mohamed~Ali Mahjoub.
\newblock Mdad: A multimodal and multiview in-vehicle driver action dataset.
\newblock In \emph{Computer Analysis of Images and Patterns: 18th International Conference, CAIP 2019, Salerno, Italy, September 3--5, 2019, Proceedings, Part I 18}, pages 518--529. Springer, 2019.

\bibitem[Liu et~al.(2022)Liu, Ren, Yu, Guo, Zhu, and Zhang]{liu2022image}
Wenyu Liu, Gaofeng Ren, Runsheng Yu, Shi Guo, Jianke Zhu, and Lei Zhang.
\newblock Image-adaptive yolo for object detection in adverse weather conditions.
\newblock In \emph{Proceedings of the AAAI Conference on Artificial Intelligence}, pages 1792--1800, 2022.

\bibitem[Mao et~al.(2021)Mao, Niu, Jiang, Liang, Chen, Liang, Li, Ye, Zhang, Li, et~al.]{mao2021one}
Jiageng Mao, Minzhe Niu, Chenhan Jiang, Hanxue Liang, Jingheng Chen, Xiaodan Liang, Yamin Li, Chaoqiang Ye, Wei Zhang, Zhenguo Li, et~al.
\newblock One million scenes for autonomous driving: Once dataset.
\newblock \emph{arXiv preprint arXiv:2106.11037}, 2021.

\bibitem[Ni et~al.(2022)Ni, Peng, Chen, Zhang, Meng, Fu, Xiang, and Ling]{ni2022expanding}
Bolin Ni, Houwen Peng, Minghao Chen, Songyang Zhang, Gaofeng Meng, Jianlong Fu, Shiming Xiang, and Haibin Ling.
\newblock Expanding language-image pretrained models for general video recognition.
\newblock In \emph{European Conference on Computer Vision}, pages 1--18. Springer, 2022.

\bibitem[Radford et~al.(2021)Radford, Kim, Hallacy, Ramesh, Goh, Agarwal, Sastry, Askell, Mishkin, Clark, et~al.]{radford2021learning}
Alec Radford, Jong~Wook Kim, Chris Hallacy, Aditya Ramesh, Gabriel Goh, Sandhini Agarwal, Girish Sastry, Amanda Askell, Pamela Mishkin, Jack Clark, et~al.
\newblock Learning transferable visual models from natural language supervision.
\newblock In \emph{International conference on machine learning}, pages 8748--8763. PMLR, 2021.

\bibitem[Schwarz et~al.(2017)Schwarz, Haurilet, Martinez, and Stiefelhagen]{schwarz2017driveahead}
Anke Schwarz, Monica Haurilet, Manuel Martinez, and Rainer Stiefelhagen.
\newblock Driveahead-a large-scale driver head pose dataset.
\newblock In \emph{Proceedings of the IEEE Conference on Computer Vision and Pattern Recognition Workshops}, pages 1--10, 2017.

\bibitem[Sun et~al.(2020)Sun, Kretzschmar, Dotiwalla, Chouard, Patnaik, Tsui, Guo, Zhou, Chai, Caine, et~al.]{sun2020scalability}
Pei Sun, Henrik Kretzschmar, Xerxes Dotiwalla, Aurelien Chouard, Vijaysai Patnaik, Paul Tsui, James Guo, Yin Zhou, Yuning Chai, Benjamin Caine, et~al.
\newblock Scalability in perception for autonomous driving: Waymo open dataset.
\newblock In \emph{Proceedings of the IEEE/CVF conference on computer vision and pattern recognition}, pages 2446--2454, 2020.

\bibitem[Tan and Le(2021)]{tan2021efficientnetv2}
Mingxing Tan and Quoc Le.
\newblock Efficientnetv2: Smaller models and faster training.
\newblock In \emph{International conference on machine learning}, pages 10096--10106. PMLR, 2021.

\bibitem[Tang et~al.(2019)Tang, Naphade, Liu, Yang, Birchfield, Wang, Kumar, Anastasiu, and Hwang]{tang2019cityflow}
Zheng Tang, Milind Naphade, Ming-Yu Liu, Xiaodong Yang, Stan Birchfield, Shuo Wang, Ratnesh Kumar, David Anastasiu, and Jenq-Neng Hwang.
\newblock Cityflow: A city-scale benchmark for multi-target multi-camera vehicle tracking and re-identification.
\newblock In \emph{Proceedings of the IEEE/CVF Conference on Computer Vision and Pattern Recognition}, pages 8797--8806, 2019.

\bibitem[Wang et~al.(2021)Wang, Yeh, and Liao]{wang2021you}
Chien-Yao Wang, I-Hau Yeh, and Hong-Yuan~Mark Liao.
\newblock You only learn one representation: Unified network for multiple tasks.
\newblock \emph{arXiv preprint arXiv:2105.04206}, 2021.

\bibitem[Wang et~al.(2023)Wang, Li, Li, Zhang, Wu, Zhong, and Sebe]{wang2023100}
Jing Wang, Wenjing Li, Fang Li, Jun Zhang, Zhongcheng Wu, Zhun Zhong, and Nicu Sebe.
\newblock 100-driver: a large-scale, diverse dataset for distracted driver classification.
\newblock \emph{IEEE Transactions on Intelligent Transportation Systems}, 2023.

\bibitem[Yin and Berger(2017)]{yin2017use}
Hang Yin and Christian Berger.
\newblock When to use what data set for your self-driving car algorithm: An overview of publicly available driving datasets.
\newblock In \emph{2017 IEEE 20th International Conference on Intelligent Transportation Systems (ITSC)}, pages 1--8. IEEE, 2017.

\bibitem[Zhao et~al.(2023)Zhao, He, Lv, Min, and Wei]{zhao2023comprehensive}
Tong Zhao, Junxiang He, Jingcheng Lv, Delei Min, and Yintao Wei.
\newblock A comprehensive implementation of road surface classification for vehicle driving assistance: Dataset, models, and deployment.
\newblock \emph{IEEE Transactions on Intelligent Transportation Systems}, 2023.

\end{thebibliography}
